\pdfoutput=1

\documentclass[11pt]{article}

\usepackage[preprint]{acl}

\usepackage{times}
\usepackage{latexsym}
\usepackage{tipa}
\usepackage{booktabs}
\usepackage{placeins}

\usepackage{multirow}

\usepackage[T2A,T1]{fontenc}

\usepackage[utf8]{inputenc}

\usepackage{microtype}

\usepackage{inconsolata}

\usepackage{graphicx}

\title{Partial Colexifications Improve Concept Embeddings}

\author{Arne Rubehn \and Johann-Mattis List \\
         Chair for Multilingual Computational Linguistics \\ University of Passau \\ Passau, Germany}

\begin{document}
\maketitle
\begin{abstract}

While the embedding of words has revolutionized the field of Natural Language Processing, the embedding of concepts has received much less attention so far. A dense and meaningful representation of concepts, however, could prove useful for several tasks in computational linguistics, especially those involving cross-linguistic data or sparse data from low resource languages. First methods that have been proposed so far embed concepts from automatically constructed colexification networks. While these approaches depart from automatically inferred polysemies, attested across a larger number of languages, they are restricted to the word level, ignoring lexical relations that would only hold for parts of the words in a given language. Building on recently introduced methods for the inference of partial colexifications, we show how they can be used to improve concept embeddings in meaningful ways. The learned embeddings are evaluated against lexical similarity ratings, recorded instances of semantic shift, and word association data. We show that in all evaluation tasks, the inclusion of partial colexifications lead to improved concept representations and better results. Our results further show that the learned embeddings are able to capture and represent different semantic relationships between concepts.

\end{abstract}

\section{Introduction}
In structural linguistics, the linguistic sign is traditionally modeled as a bipartite structure, consisting of a \emph{form} that signifies a \emph{meaning} \citep{saussure1916course}. 
Word embeddings computed from large text collections \citep{mikolov2013efficient} offer a convenient way to express semantic properties of words in a particular language in low-dimensional spaces. In such an approach, meaning is defined only structurally, in relation to other meanings. Whether word embeddings capture meaning in the form of a
``relation between a linguistic form and
communicative intent'' \citep{Bender2020}, however, can at least be doubted.
 
While word embeddings provide information on semantic relations inside individual languages, \emph{cross-linguistic colexifications} -- collections of senses that are expressed by a single word form across multiple languages (compare German \emph{böse} colexifying the senses `evil' and `angry') -- capture a different kind of semantic relation. Only small amounts of colexifications can be observed for any individual language in the world, yet taking the languages of the world together, we can infer dense concept networks from cross-linguistic colexification patterns that show fascinating semantic structures \citep{List2013a}. 
In order to employ these patterns in machine learning approaches, previous studies have proposed to embed the underlying graphs, using graph embedding techniques by which nodes in a graph are projected on vector spaces that represent their relation to other nodes in the graph \citep{harvill2022syn2vec}. Given that colexification networks represent concepts defined independently of individual languages, concept embeddings can therefore be learned from colexification networks in a similar way that word embeddings are learned from distributional data.


All concept embeddings methods presented so far have exclusively relied on \emph{full colexifications}. This means that the concept networks were derived from those cases in which two or more senses are colexified by the \emph{same} word form in a given language. \emph{Partial colexifications} -- colexification patterns in which words do not need to match entirely in order to colexify two or more senses --, on the other hand, have not been tested so far. 

Here, we present initial tests in which both full and partial colexification data are used to train concept embeddings with the help of different graph embedding techniques. We evaluate the quality of the learned embeddings on three different tasks, including the modeling of semantic similarity, the prediction of semantic change, and the prediction of word associations. Our results show that enriching concept embeddings with partial colexification data greatly improves their performance on all three tasks, outperforming similarity metrics inferred from the graph topology directly, and surpassing static word embeddings in two out of three tasks.

\section{Background}

\subsection{Word Embeddings}

Based on the distributional hypothesis \citep{harris1954distributional,firth1957synopsis}, word embeddings are dense, low-dimensional representations of words learned from their co-occurrence distributions. With their ability to capture different aspects of semantics, they have laid the groundwork for the success of state-of-the-art techniques in language processing. Static embedding models like Word2Vec \citep{mikolov2013efficient}, GloVE \citep{pennington2014glove}, and fastText \citep{bojanowski2017enriching} have quickly gained popularity and are still widely used for modeling lexical semantics in a latent space, mainly because they are more transparent and interpretable than contextual embeddings employed in Transformer architectures \citep{vaswani2017attention} such as BERT \citep{devlin2018bert} and GPT \citep{radford2019language}.

Word embeddings are inherently monolingual and can therefore not be used directly in cross-linguistic applications. In order to address this problem, multiple approaches for obtaining multilingual embeddings have been proposed. While some approaches map monolingual embeddings onto a shared space \citep{ammar2016massively,artetxe2018robust,lample2018word}, other approaches learn multilingual embeddings directly from parallel corpora \citep{dufter2018embedding,levy2016strong}.



\subsection{Colexification Networks}\label{sec:colex}
Colexification, originally introduced by \citet{Francois2008}, is a cover term for both polysemy and homophony. Thinking of the word forms in a given language being linked to different \emph{senses} or \emph{concepts}, two senses are thought to be colexified when there is one word form that expresses both of them at the same time. While it would be difficult to assess \emph{why} two words colexify, given that it is not a trivial task to distinguish polysemy from homophony \citep{Tjuka2023}, it is computationally very easy to \emph{infer} colexifications from comparative wordlists where a list of concept is translated into one or more languages \citep{List2022COL}.
Collections of colexification across multiple languages can modeled in the form of \emph{colexification networks} \citep{List2013a}, revealing interesting colexification patterns that may reflect typological or areal trends in lexical typology \citep{List2018e}. The \emph{Database of Cross-Linguistic Colexifications} (CLICS, \citealt{rzymski2020clics}, \href{https://clics.clld.org}{https://clics.clld.org}) offers a large collection of colexifications inferred from several thousand languages that can be used in computational approaches or interactively inspected. 

Given that all semantic change takes polysemy as its starting point \citep{Traugott2002}, colexification networks prove useful when investigating common pathways and tendencies of semantic change \citep{munch2015evaluating,dellert2024causal,Bocklage2024PREPRINT}. Given that they also reflect certain aspects of conceptual proximity, they have been used to study semantic similarity in different lexical domains, including emotion semantics \citep{Jackson2019}, body part terminology \citep{Tjuka2024a}, and affective meaning \citep{dinatale2021colexification}.

While most colexification studies only take \emph{full} colexifications into account, where identical word forms colexify different senses, \citet{Urban2011} has shown that semantic relations derived from word parts -- reflecting overt marking in lexical coding -- may also reveal interesting structures that provide additional hints on the directionality of semantic change in particular and lexical motivation in general. \citet{List2023a} applies this idea to a smaller dataset of 329 languages taken from the \emph{Intercontinental Dictionary Series} \citep{key2016ids}, showing that \emph{partial colexifications} can be inferred conveniently in two flavors. \emph{Affix colexifications} point to directional relations between two concepts where one of the words expressing the concepts is a prefix or a suffix -- in the computational sense -- of the other word. \emph{Overlap colexifications} point to undirected relations between concepts that share a substring of a given length. 
Overlap colexifications usually connect morphologically complex words that are derived or compounded using the same stem, while affix colexifications directly connect derived or compounded forms to their stems.
Examples for full and partial colexifications according to this notion are given in Table \ref{tab:1}.  

\begin{table*}[htb]
\centering
\resizebox{\textwidth}{!}{
\tabular{llllll}
\toprule
\bfseries Language & \bfseries Word A & \bfseries Concept A & \bfseries Word B & \bfseries Concept B & \bfseries Colexification\\\midrule
Yaqui & \textipa{dZ u j a} & ``tree'' & \textipa{dZ u j a} & ``forest'' & full \\
Guìlín (China) & \textipa{\textctc y} \textsuperscript{21} &``tree'' & \textipa{\textctc y} \textsuperscript{21} \textipa{l i N} \textsuperscript{21} & ``forest'' & affix \\
Fúzhōu (China) & \textipa{ts\textsuperscript{h} j eu} \textsuperscript{212} \textipa{p\textsuperscript{h} w oi}\textsuperscript{5} & ``bark'' & \textipa{ts\textsuperscript{h} j eu} \textsuperscript{212} \textipa{l i N} \textsuperscript{53} & ``forest'' & overlap \\\bottomrule
\endtabular}
\caption{Different colexification types, following \citet{List2023a} with examples across different languages (superscript letters code for tones in Sinitic examples).}
\label{tab:1}
\end{table*}

\subsection{Graph Embeddings}

Inspired by the success of word embeddings and deep neural networks, graph embedding techniques have gained traction over the past decade. The task of embedding a graph is -- similar to word embeddings -- to learn low-dimensional representations for each node in a graph, such that properties of and relationships between nodes are preserved. Representing nodes in an embedded space has several advantages over the direct representation of a graph as a set of nodes and edges: The computational complexity decreases, downstream methods can be parallelized, and nodes can directly serve as inputs for machine learning models \citep{cui2019survey}.

Most models for learning graph embeddings can be categorized according to three basic strategies, where we find models based on \textit{matrix factorization} \citep{cao2015grarep,ou2016asymmetric,zhang2019prone}, \textit{random walks} \citep{perozzi2014deepwalk,grover2016node2vec}, and \textit{deep learning} \citep{wang2016sdne,kipf2016variational,hamilton2017inductive}. Common applications of graph embeddings include node classification, link prediction, and network visualization \citep{cui2019survey,goyal2018graph}.


\subsection{Concept Embeddings}

Compared to the popularity of word embeddings, the embedding of concepts or word senses have received only little attention so far. For monolingual word senses, WordNet \citep{fellbaum1998wordnet} is often used as a reference catalogue whose synsets are embedded, usually for word sense disambiguation purposes \citep{trask2015sense2vec,rothe2015autoextend,jafarinejad2023synset2node}. Closest to our work are studies that learn concept embeddings from multilingual data. \citet{harvill2022syn2vec} learn concept embeddings from BabelNet \citep{navigli2012babelnet}, an automatically crawled multilingual extension of WordNet. \citet{chen2023colex2lang} combine BabelNet and CLICS to learn concept embeddings, which are aggregated to represent languages based on their colexification patterns. \citet{conia2020conception} infer vector representations from BabelNet as well, however those are sparse and high-dimensional. Finally, \citet{liu2023crosslingual} train concept embeddings from colexification networks that are inferred automatically over parallel multilingual text corpora.

We intend to improve on recent work in two ways. First, our embeddings are trained on manually curated colexification data, which is less error-prone than automatically constructed colexification graphs. Second, we do not only take full colexification into account, but also consider partial colexifications. We believe that this adds complementary information of which the potential has not been explored yet.

\section{Materials and Methods}

\subsection{Colexification Data}
We use colexification data inferred by \citet{List2023a} from the \textit{Intercontinental Dictionary Series} \citep[IDS, \href{https://ids.clld.org}{https://ids.clld.org}, ][]{key2016ids} spanning 329 language varieties and 1,310 concepts. The inventory of concepts is defined by Concepticon \citep[\href{https://concepticon.clld.org}{https://concepticon.clld.org}, Version 3.3, ][]{concepticon}, a reference catalog that defines more than 4,000 different senses (referred to as \emph{concept sets}), taken from questionnaires, wordlists, or concept lists that are typically used for language documentation. Being built upon resources designed by linguists to document languages all over the world, concept sets in Concepticon are defined independently of the languages in which they are expressed, even if the language used to gloss these concepts is typically English. This means that concept sets defined by Concepticon represent a cross-linguistic resource that attempts to define concepts that are expressed coherently across many of the world's languages.

The colexification data from \citet{List2023a} comprises not only \textit{full} colexifications, but also \textit{affix} and \textit{overlap} colexifications, as described above (see §~\ref{sec:colex}).
From the colexifications identified in \citet{List2023a}, it is straightforward to build three colexification networks, one for each colexification type. In each network, the concept sets form the nodes that are connected by edges weighted by language families: If two concepts are connected with a weight of 5, this indicates that at least one instance of this colexification can be found in five different language families. Full and overlap colexifications are undirected by nature and thus form undirected graphs, whereas affix colexifications are inherently directed. However, since all downstream tasks only measure symmetric similarities between concepts, and most graph embedding techniques cannot account for directionality \citep{cui2019survey}, the affix colexification network is transformed to an undirected network as well. An overview of the three resulting graphs is given in Table \ref{tab:colex-stats}.

\begin{table}[b]
\centering
\resizebox{\linewidth}{!}{%
\begin{tabular}{@{}lll@{}}
\toprule
\bfseries Colexification Type &\bfseries  Nodes (Concept Sets) & \bfseries Edges  \\ \midrule
full                & 1,246 & 4,008  \\
affix               & 1,308 & 38,215 \\
overlap             & 926   & 12,974 \\ \bottomrule
\end{tabular}}
\caption{Number of nodes (concept sets) and edges in the three colexification networks used in our study.}
\label{tab:colex-stats}
\end{table}

\subsection{Concept Embeddings}

We employ and compare three different graph embedding techniques to learn low-dimensional representations of the nodes (i.e. the concepts) in each graph; each method representing one of the three main paradigms for graph embedding techniques. SDNE \citep[\emph{Structural Deep Network Embedding}, ][]{wang2016sdne} is a deep autoencoder that optimizes embeddings towards retaining first- and second-degree neighborhoods. Node2Vec \citep{grover2016node2vec} is a well-known graph embedding technique that samples random walks from the graph, which then are used as training data for a simple Word2Vec model \citep{mikolov2013efficient}. Due to the sampling of random walks, this technique is able to learn some more distant relationships between nodes, while still preserving close community structures. Finally, ProNE \citep{zhang2019prone} employs sparse matrix factorization enhanced by spectral propagation, leading to extremely efficient training of high-quality embeddings.

For each of the three colexification networks and with each of the three embedding techniques, 128-dimensional embeddings are trained with fairly standard hyperparameters. In a postprocessing step, the embeddings trained on the full colexification network are combined with embeddings for either or both of affix and overlap colexifications. Following \citet{harvill2022syn2vec}, this is achieved by concatenating the embeddings and then employing Principal Component Analysis (PCA) to reduce them back to the original dimensionality. With this, six sets of embeddings are obtained per embedding technique, combining information from different types of colexification: \textit{full}, \textit{affix}, \textit{overlap}, \textit{full/affix}, \textit{full/overlap}, and \textit{full/affix/overlap}.

\subsection{Baselines}

The quality of the trained embeddings is evaluated against four simple methods to infer similarities (or distances) between nodes from the graph directly. The first baseline metric is the length of the \textit{shortest path} between two nodes on a graph with inverted weights, where each weight $w$ is transformed by the simple function $f(w) = w^{-1}$. This is necessary since the weights in the original graphs indicate similarities rather than distances, with higher weights between nodes indicating a stronger connection.

In the remaining three similarity metrics, we follow \citet{DeDeyne2018} who compare \textit{cosine similarity}, \textit{positive pointwise mutual information (PPMI)}, and \textit{random walk similarity} on word association networks. The first two metrics can be easily inferred from the graph's adjacency matrix by calculating the cosine similarity between the rows of individual concepts, or the PPMI respectively. The last metric simulates random walks from each node and measures how similar the average random walks for different start nodes are.

Finally, we map the embedded concepts to fastText embeddings in nine different languages \citep{grave2018learning} via parallel word pairs in Multi-SimLex \citep{vulic2020multi}, which has been manually linked to Concepticon concept sets in the past \citep{List2021MS}. For concepts with multiple lexical realizations (e.g. the concept set \textsc{car} corresponds to Russian \textit{avtomobil'}  
and 
\textit{mashina}, two words that are both typically used to refer to a car in Russian),
the fastText vector corresponding to that concept is the mean of the vectors of the respective words, weighted by their relative frequency. For a fair comparison, we reduce the dimensionality of the vectors from originally 300 to 128 components, the same size as the trained concept embeddings. We report the averaged and the best performance of these embeddings in all available languages and provide a direct comparison to the trained concept embeddings.

\subsection{Experiments}

We conduct three experiments, in which we test how well our concept embeddings align with different aspects of semantic similarity, concentrating on human similarity ratings for individual word pairs across multiple languages (\emph{Modeling Lexical Semantic Similarity Task}, §~\ref{sec:e1}), semantic change processes as documented in the linguistic literature (\emph{Predicting Semantic Change Task}, §~\ref{sec:e2}), and word associations derived from experimental data (\emph{Predicting Word Associations Task}, §~\ref{sec:e3}). With these three tests, we intend to cover three scenarios in which concept embeddings might play an important role in future research.
All three evaluations are based on similarities between concept pairs, inferred by calculating the cosine similarity between embedding vectors of individual concept sets.

\subsubsection{Modeling Lexical Semantic Similarity}\label{sec:e1}

The \textit{lexical semantic similarity} task \citep[LSIM, ][]{harvill2022syn2vec} measures to which degree the similarities between embeddings correlate with human similarity judgements. We obtain these ratings from Multi-SimLex \citep{vulic2020multi}, which encompasses similarity judgments for 1,888 word pairs in 12 typologically diverse languages. 546 of these pairs have been mapped to Concepticon \citep{List2021MS}, excluding word pairs for which no counter-part could be found in Concepticon. We further exclude near-synonyms linking to the same Concepticon concept sets, resulting in 538 concept pairs. The inferred embedding similarities are evaluated against this subsample. For each concept pair, we calculate the average similarity rating over all languages in Multi-SimLex in order retrieve a cross-linguistic notion of semantic similarity that accounts for outliers in the data. The rating for the concept pair ``bank'' vs. ``seat'', for example, has a much higher similarity rating in Spanish than in any other language, due to the fact that Spanish \textit{banco} is polysemous, referring to both ``bank'' and ``bench''. Following previous studies \citep{conia2020conception,harvill2022syn2vec} we calculate Spearman's rank correlation \citep{Spearman1904} between the averaged multilingual similarity ratings and the cosine similarities retrieved from the concept embeddings for different models and settings.

\subsubsection{Predicting Semantic Change}\label{sec:e2}
Given the importance of polysemy and colexication data for the handling of semantic change patterns, we are interested to know how well concept embeddings can distinguish likely patterns of semantic change from less likely ones. 
To test this, we extracted several hundred historically attested semantic shifts from the \emph{Database of Semantic Shifts} (DatSemShift, \href{https://datsemshift.ru/}{https://datsemshift.ru/}, \citealt{Zalizniak2024}) a database that collects and structures individual instances of semantic change from the literature. While the database offers various data on semantic change, ranging from colexifications, via partial colexifications, up to cognate words with diverging meanings, we extracted 547 concept pairs in which the shift took place from an attested ancestral language to an attested daughter language, thus guaranteeing that these shifts can be observed directly through the comparison of sources. For the extraction, we used a computer-readable version of the database \citep{Bocklage2024TBLOG04} that codes the data in Cross-Linguistic Data Formats \citep[CLDF, \href{https://cldf.clld.org}{https://cldf.clld.org}, ][]{Forkel2018a}, a format specification for linguistic data that facilitates the use in computational applications.
We frame semantic change prediction as a binary classification task, where the actual shifts are instances of the positive class, while the instances of the negative class are constructed by replacing one of the concepts in each pair by a randomly selected concept. This process is known as \textit{negative sampling} and is often employed in training objectives to prevent overfitting of models \citep{smith2005contrastive,gutmann2010noise,mikolov2013distributed}. We train a simple logistic regression classifier on this sample with the embedding similarity of the respective concept pair as sole input feature. To even out coincidental biases in the negative sampling, we run the sampling procedure 50 times and fit a logistic regression on the sample. The negative samples are shared across all models and baselines to ensure a fair comparison. We report the prediction accuracy averaged over all samples for each model.

\subsubsection{Predicting Word Associations}\label{sec:e3}

\textit{Link prediction} is one of the central problems in network analytics, where it is mainly used to recover missing data in sparse networks or predict likely links that might arise in the future \citep[e.g. predicting which authors are likely to collaborate in the future based on current co-authorship networks; ][]{libennowell2007link}. For evaluation purposes, this task is usually framed in a way that links from a network are predicted using information from another network in the same domain (such that the nodes of both networks can easily mapped onto each other). Following this paradigm, we use embedding similarities in order to predict links from the \emph{Edinburgh Association Thesaurus} \citep[EAT, ][]{Kiss1973}, a large-scale network constructed from data in word association studies. In the experiment, participants were given cue words to which they were to respond with the first word that came to their mind. As \citet{Tjuka2022} show with their \emph{Database of Norms, Ratings, and Relations} (NoRaRe, \href{https://norare.clld.org}{https://norare.clld.org}), a network from the EAT data can be easily inferred by counting all cue-response pairs as weighted edges and linking them automatically to the concept sets in Concepticon. NoRaRE assigns 1,787 of the 55,732 nodes in the EAT to concept sets in Concepticon. In order to filter out noise, we remove edges with a weight of less than five. Reducing the association network further to concepts present in the colexification data by \citet{List2023a} -- the concept space that is being embedded -- results in a network of 746 concepts with 780 edges. We formalize the link prediction task as a binary classification task with negative sampling; conceptually identical to the prediction of semantic change. Again, we draw 50 negative samples and report the average accuracy.
\begin{table*}[!t]
\centering
\resizebox{0.75\textwidth}{!}{%
\tabular{c}
\begin{tabular}{@{}lllllll@{}}
\toprule
\multirow{2}*{\bfseries Method} & \multicolumn{6}{c}{\bfseries Colexification Type} \\ 
                  & \itshape\bfseries full  & \itshape\bfseries affix & \itshape\bfseries overlap & \itshape\bfseries full / affix & \itshape\bfseries full / overlap & \itshape\bfseries full / affix / overlap \\ \midrule
shortest path      & -0.60 & -0.49 & -0.18 & -0.51 & -0.45 & -0.47            \\
cosine similarity  & 0.50 & 0.42 & 0.15 & 0.50 & 0.27 & 0.39               \\
PPMI               &   0.53 & 0.33 & 0.21 & 0.31 & 0.34 & 0.23               \\
random walks       &  0.43 & 0.34 & 0.10 & 0.38 & 0.22 & 0.25             \\ \midrule
SDNE               &   0.48 & 0.36 & 0.36 & 0.60 & 0.46 & 0.47            \\
Node2Vec           &  0.64 & 0.58 & 0.26 & \underline{0.69} & 0.61 & 0.66              \\
ProNE              &    0.64 & 0.63 & 0.26 & \textbf{0.72} & 0.62 & 0.66            \\ \midrule
fastText-ZH (best) & \multicolumn{6}{c}{0.44}                                                     \\
fastText (avg.)    & \multicolumn{6}{c}{0.36}                                                     \\ \bottomrule 
\end{tabular}
\\
(a) Spearman rank correlations for Multi-SimLex similarity ratings and concept similarities.
\\\\
\begin{tabular}{@{}lllllll@{}}
\multirow{2}*{\bfseries Method} & \multicolumn{6}{c}{\bfseries Colexification Type} \\ 
                  & \itshape\bfseries full  & \itshape\bfseries affix & \itshape\bfseries overlap & \itshape\bfseries full / affix & \itshape\bfseries full / overlap & \itshape\bfseries full / affix / overlap \\ \midrule
shortest path      & 0.79 & 0.73 & 0.69 & 0.78 & 0.79 & 0.79    \\
cosine similarity  & 0.69 & 0.74 & 0.65 & 0.76 & 0.70 & 0.75 \\
PPMI               &  0.71 & 0.73 & 0.62 & 0.78 & 0.73 & 0.79    \\
random walks       &  0.71 & 0.72 & 0.65 & 0.74 & 0.73 & 0.74  \\ \midrule
SDNE               &  0.66 & 0.66 & 0.70 & 0.78 & 0.75 & 0.75 \\
Node2Vec           &  0.79 & 0.75 & 0.72 & \textbf{0.83} & 0.82 & \textbf{0.83}   \\
ProNE              &  0.78 & 0.78 & 0.71 & \underline{0.82} & 0.80 & \textbf{0.83}         \\ \midrule
fastText-ET (best) & \multicolumn{6}{c}{\underline{0.82}}                                                     \\
fastText (avg.)    & \multicolumn{6}{c}{0.78}                                         \\ \bottomrule 
\end{tabular}
\\
(b) Predicting semantic changes from DatSemShift.
\\
\\
\begin{tabular}{@{}lllllll@{}}
\toprule
\multirow{2}*{\bfseries Method} & \multicolumn{6}{c}{\bfseries Colexification Type} \\ 
                  & \itshape\bfseries full  & \itshape\bfseries affix & \itshape\bfseries overlap & \itshape\bfseries full / affix & \itshape\bfseries full / overlap & \itshape\bfseries full / affix / overlap \\ \midrule
shortest path      & 0.71 & 0.76 & 0.70 & 0.77 & 0.67 & 0.78     \\
cosine similarity  &0.64 & 0.73 & 0.65 & 0.74 & 0.65 & 0.73           \\
PPMI               & 0.65 & 0.76 & 0.68 & 0.77 & 0.67 & 0.78      \\
random walks       & 0.65 & 0.70 & 0.65 & 0.71 & 0.67 & 0.71   \\ \midrule
SDNE               & 0.59 & 0.67 & 0.68 & 0.72 & 0.72 & 0.74  \\
Node2Vec           & 0.71 & 0.75 & 0.72 & 0.78 & 0.77 & 0.79    \\
ProNE              & 0.71 & 0.79 & 0.71 & 0.80 & 0.77 & \underline{0.81}           \\ \midrule
fastText-EN (best) & \multicolumn{6}{c}{\textbf{0.87}}                                                     \\
fastText (avg.)    & \multicolumn{6}{c}{\underline{0.81}}                                                     \\ \bottomrule 
\end{tabular}
\\
(c) Accuracies of link prediction in the EAT.
\endtabular
}
\caption{Performance of concept embeddings in various flavors across our three evaluation tasks. Best models are marked in \textbf{bold} font. Second best models are marked by \underline{underlining} the value.}
\label{tab:results}
\end{table*}

\subsection{Implementation}
All training and evaluation procedures were implemented in Python 3.12. The training routine for SDNE and Node2Vec was implemented from scratch using PyTorch \citep{paszke2019pytorch} and Scikit-learn \citep{pedegrosa2011sklearn}. The ProNE models were trained using the {NodeVectors} package \citep{ranger2024nodevectors}. The shortest path baseline was computed using the NetworkX \citep{hagberg2008exploring} implementation of Dijkstra's algorithm \citep{dijkstra1959note}. Spearman rank correlations were computed using SciPy \citep{Virtanen2020}. Evaluation data was retrieved using {PyCLDF} \citep{PyCLDF}, {PyConcepticon} \citep{PyConcepticon}, and {PyNoRaRe} \citep{PyNoRaRe}. Visualizations were created using {Matplotlib} \citep{Hunter2007} and {adjustText} \citep{flyamer2024adjusttext} packages. The supplementary material offers access to all data and code needed to replicate the analyses described in this study, including a detailed account of the package versions that were used in our implementation.

\section{Results}

Table \ref{tab:results} summarizes the performance of the different models and baselines on the three evaluation tasks. Since fastText embeddings are trained independently from the different colexification networks, the reported metric does not relate to the different types of colexification. As indicated in the previous section, we report the performance of the best language and the average performance across languages in fastText.

Some general patterns can be observed across all three tasks. Concerning the baselines -- similarities (or distances for the \textit{shortest path} metric) inferred directly from the graph -- simply measuring the length of the shortest path consistently is a better predictor than the three methods discussed in \citet{DeDeyne2018}. This comes at the expense that path lengths can only be inferred between connected nodes; however, this problem can easily be solved by assigning a default value -- the similarity between disconnected nodes in all other three baselines is always 0.

Regarding the models, it is clear that SDNE is not viable for embedding colexification graphs, as it performs substantially worse than the other two embedding methods, and is even beaten by simpler baselines in most experiments. This model therefore will not be discussed in detail, instead, we will focus on the more performant models, Node2Vec and ProNE. These two models perform almost on par, with only a few differences, the most notable being that ProNE consistently outperforms Node2Vec on affix colexifications.

The latter two tasks, semantic change prediction (Table \ref{tab:results}b) and association link prediction (Table \ref{tab:results}c) show the same trends and mostly mirror each other. Since the EAT is not a multilingual, but a monolingual evaluation resource in English, it is not surprising that the English word embeddings are the best predictor for English word associations. However, word embeddings from other languages are generally good at capturing associative links -- even across languages -- and on average perform similar to the best concept embedding model (ProNE, all colexifications). For the concept embeddings, it is clearly discernible that including partial colexifications improves the embedding quality in both prediction tasks. This holds true for both affix and overlap colexifications. Node2Vec and ProNE both prove themselves as reliable graph embedding techniques with very similar performances; Node2Vec has a slight edge over ProNE in predicting semantic shifts, while ProNE performs slightly better on the EAT data. Both models outperform fastText embeddings when trained on the entire colexification data.

\begin{figure*}[t]
    \begin{tabular}{cc}
        \includegraphics[width=\columnwidth]{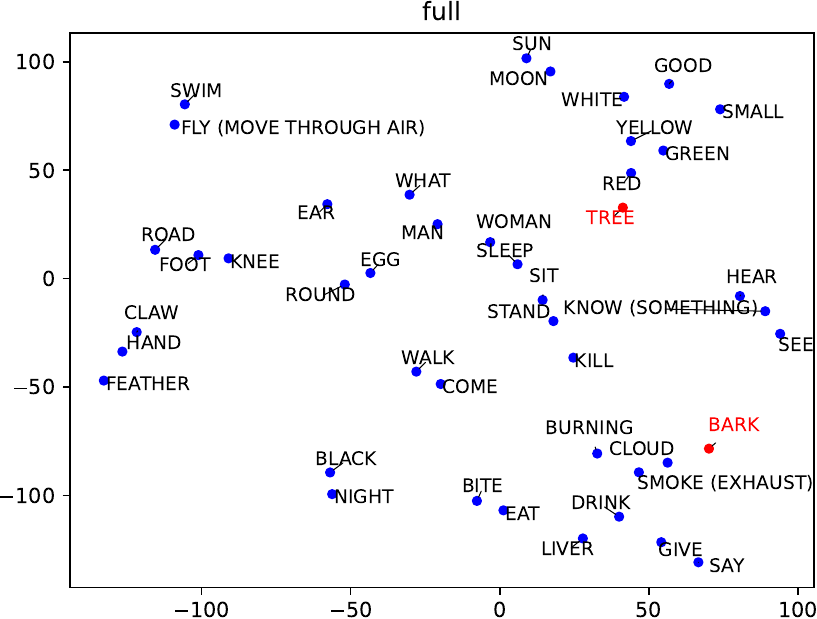} & \includegraphics[width=\columnwidth]{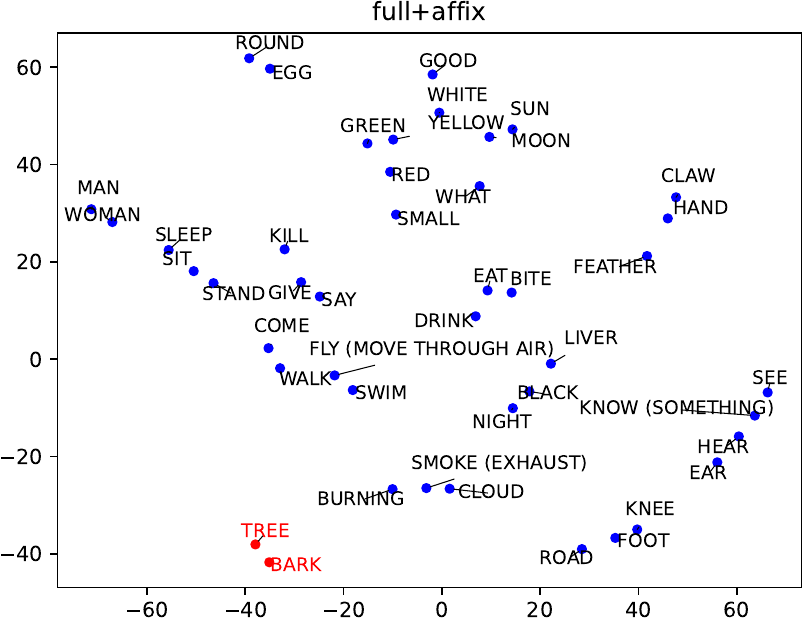} \\
    \end{tabular}
    \caption{Two-dimensional visualizations of embeddings only learned on full colexification data (left) and embeddings learned on full and affix colexifications (right), created using t-SNE, for a small list of concepts taken from the Swadesh list of 100 items \citep{Swadesh1955}.}
    \label{fig:tsne}
\end{figure*}

The lexical semantic similarity (LSIM, Table \ref{tab:1}a) task shows some different trends. Overlap colexifications correlate poorly with perceived lexical similarity, and the inclusion of overlap colexifications is consistently detrimental to the embeddings' performance on LSIM. On the other hand, the combination of full and affix colexifications is very fruitful: Across all embedding models, this set-up yields the best performance by far. Again, Node2Vec and ProNE behave very similarly in terms of overall performance, but the latter outperforms the former in embedding affix colexifications, which propagates through the combined embeddings.

Word embeddings perform notably poorly on LSIM, with significantly lower correlations than simple baseline metrics inferred directly from full colexification networks. The weak correlation with perceived lexical similarity and the high accuracies on predicting word associations provide further support for the claim that word embeddings capture semantic association rather than semantic similarity \citep{yang2022lexical,karidi2024aligning}.

The results show that concept embeddings generally benefit from partial colexification data. In both prediction tasks, embeddings learned on full colexifications alone perform significantly worse than those enhanced by either or both types of partial colexification. However, overlap colexification patterns are rather detrimental than helpful for assessing semantic similarity. Integrating affix colexifications, on the other hand, is clearly beneficial for all three downstream tasks, including LSIM.

\section{Discussion and Outlook}

Our results underline the benefits of combining full and partial colexification patterns when inferring concept embeddings from cross-linguistic colexification data. The combined embeddings consistently outperform embeddings based on full colexifications in our three different downstream tasks. It seems that partial colexifications -- in their two flavors of affix colexifications and overlap colexifications, as tested in our study -- can capture conceptual relationships that are rarely expressed by full colexifications alone. As an example, consider Figure \ref{fig:tsne}, where 44 concepts from Swadesh's list of 100 concepts \citep{Swadesh1955} are plotted along two dimensions based on their embeddings learned from both full colexifications (left) and full colexifications combined with affix colexifications (right, created using t-SNE). 
While \textsc{tree} and \textsc{bark} are far apart in the embedded space learned from full colexifications alone, they appear very close to each other when affix colexifications are considered as well. This nicely reflects that words for \textsc{bark} are often compounds that contain \textsc{tree} as a member, indicating a clear conceptual relation, although the two concepts are rarely fully colexified.

So far, related studies have only considered full colexifications that have been inferred from multilingual corpus data \citep{liu2023crosslingual,harvill2022syn2vec,chen2023colex2lang}. Our findings imply that corpus-based approaches could benefit from looking into partial colexifications in order to improve the representation of semantic relations across multiple languages. However, inferring morphological structures from corpus data still poses a challenge to which a widely applicable solution has not yet been found \citep{manova2020what}.
 
Since our embeddings are based on colexifications attested in multilingual wordlists rather than being automatically inferred from texts they suffer less from noise introduced during colexification inference, resulting in higher-quality concept embeddings. The drawback of this approach is, however, a lower coverage resulting from smaller networks. With growing amounts of standardized multilingual wordlists \citep{list2022lexibank}, we are confident that the conceptual space covered by colexification networks -- as well as by the resulting embeddings -- will increase in the future. Being able to model meaning independently of individual languages in a continuous space could advance computational approaches to historical linguistics and linguistic typology in many ways, providing new possibilities for automatic cognate detection \citep{wu2020computer} and the reconstruction of semantic change scenarios \citep{Urban2015}.



\section*{Supplementary Material}

All data and code underlying this study are available from the supplementary material accompanying this paper. They are curated on GitHub (\url{https://github.com/calc-project/concept-embeddings}, \nolinebreak v.0.1) and archived with Zenodo (\url{https://doi.org/10.5281/zenodo.14866618}).

\section*{Limitations}

The major and most obvious limitation of the present approach is that it currently only covers around 1,000 core concepts, due to the fact that the underlying colexification networks are mostly inferred from comparative wordlists for basic vocabulary. While the restriction to basic concepts is a common practice in historical linguistics and typology, it poses an obstacle to direct applications on multilingual tasks in NLP.

The graph embedding techniques discussed in this paper rely on the graph topology alone: disconnected nodes (concepts with no attested colexifications), therefore, cannot be embedded with this approach. To circumvent this problem of missing data in downstream applications, future approaches might benefit from using node features (e.g. part-of-speech) on top of the graph topology. Recent advances in Graph Neural Networks have sucessfully shown the potential of combining these two pieces of information \citep[e.g.][]{hamilton2017inductive}.

Finally, while our results clearly show that affix colexifications provide crucial information complementary to what is encoded in a full colexification network, the role of overlap colexifications remains unclear. Further research is required to better understand what kind of semantic and conceptual relations different types of colexification represent.

\section*{Acknowledgments}

We would like to thank Yannick Funk for his valuable advise on implementing the graph embedding models in PyTorch. This project was supported by the ERC Consolidator Grant ProduSemy (Grant No. 101044282, see \url{https://doi.org/10.3030/101044282}). Views and opinions expressed are however those of the author(s) only and do not necessarily reflect those of the European Union or the European Research Council Executive Agency (nor any other funding agencies involved). Neither the European Union nor the granting authority can be held responsible for them.

\bibliography{custom}
\newpage
\appendix
\onecolumn
\section{Visualizations}
\begin{figure*}[!h]
\resizebox{\textwidth}{!}{
    \begin{tabular}[!htb]{cc}
        \includegraphics[width=\columnwidth]{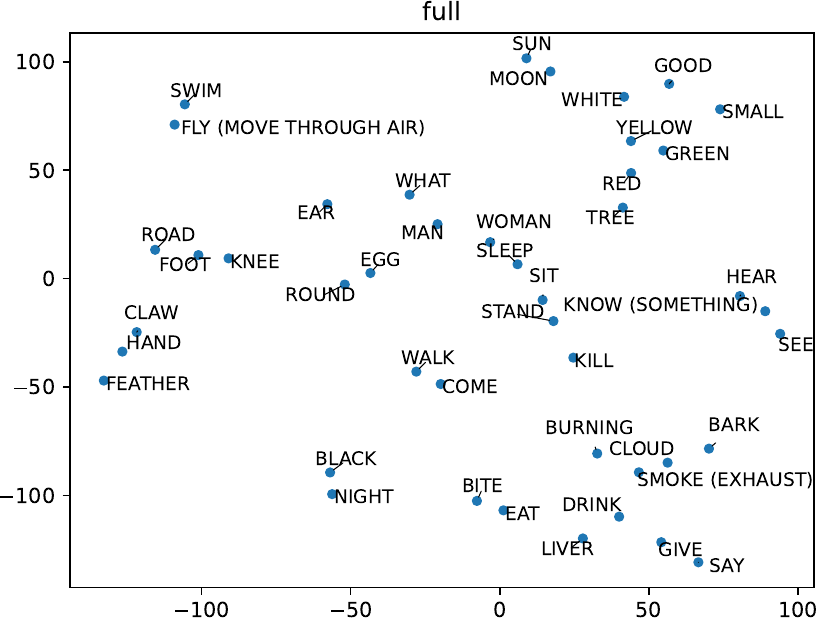} & \includegraphics[width=\columnwidth]{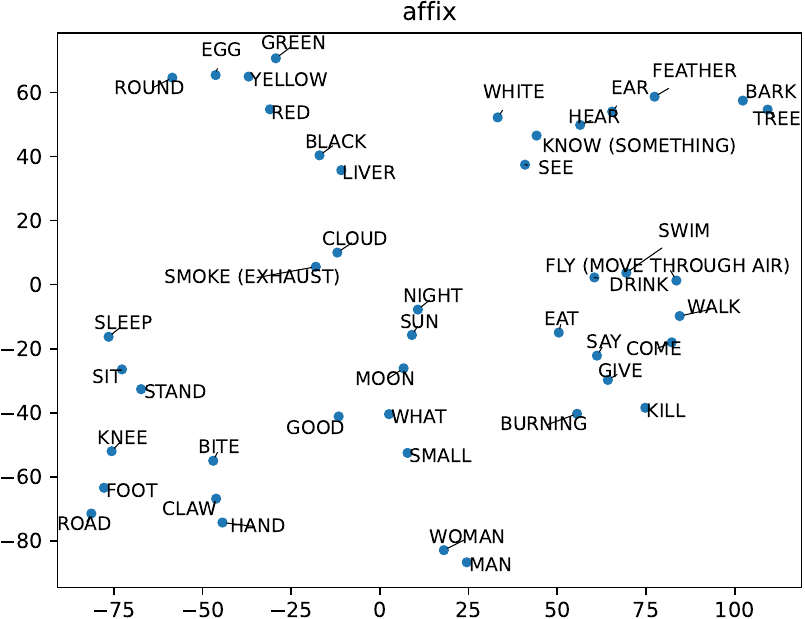} \\
        \includegraphics[width=\columnwidth]{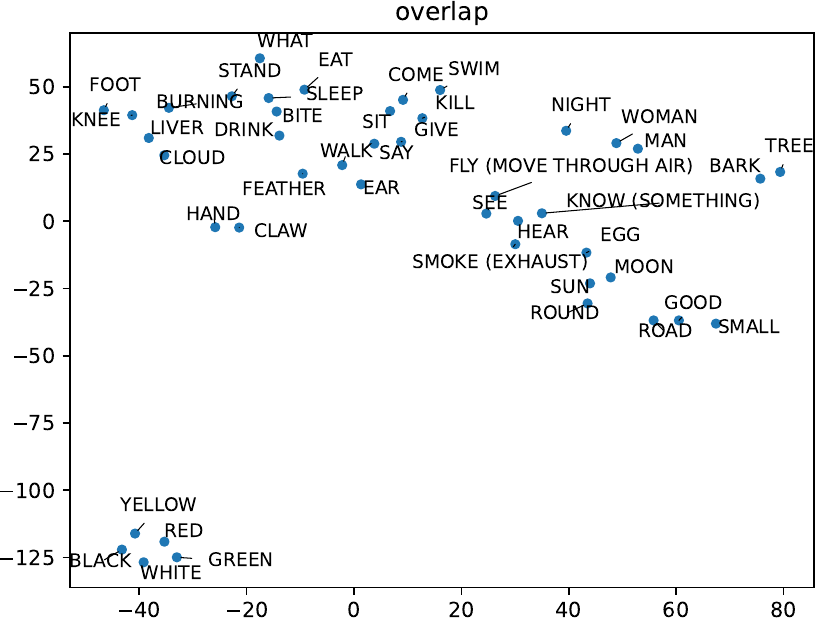} & \includegraphics[width=\columnwidth]{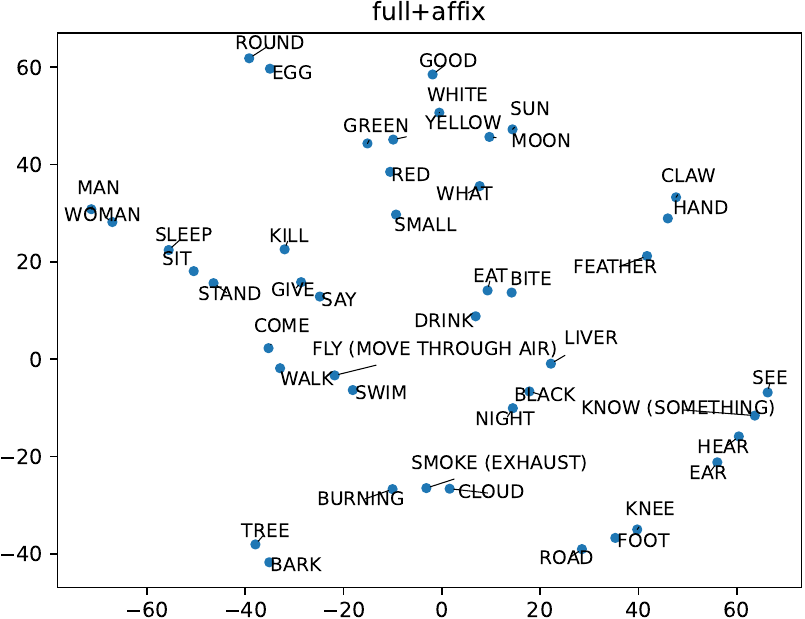} \\
        \includegraphics[width=\columnwidth]{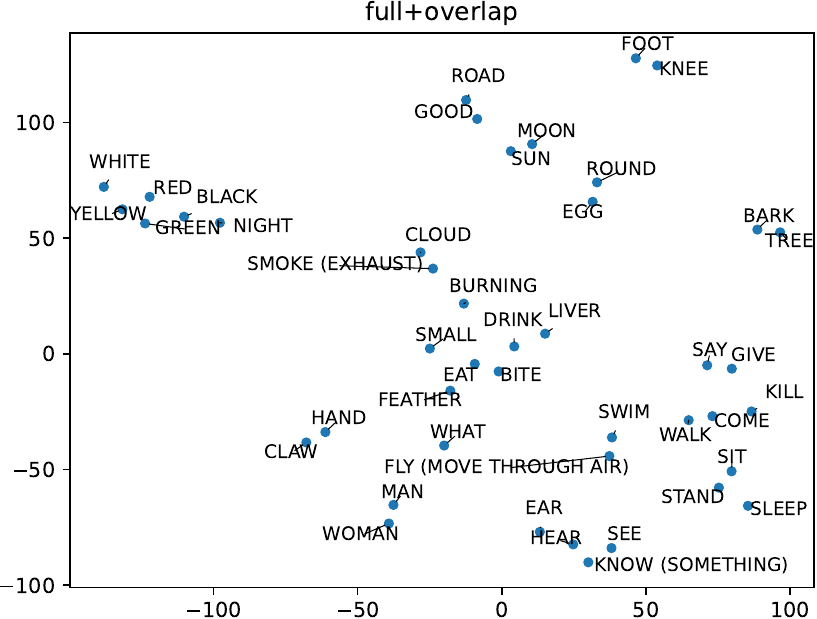} & \includegraphics[width=\columnwidth]{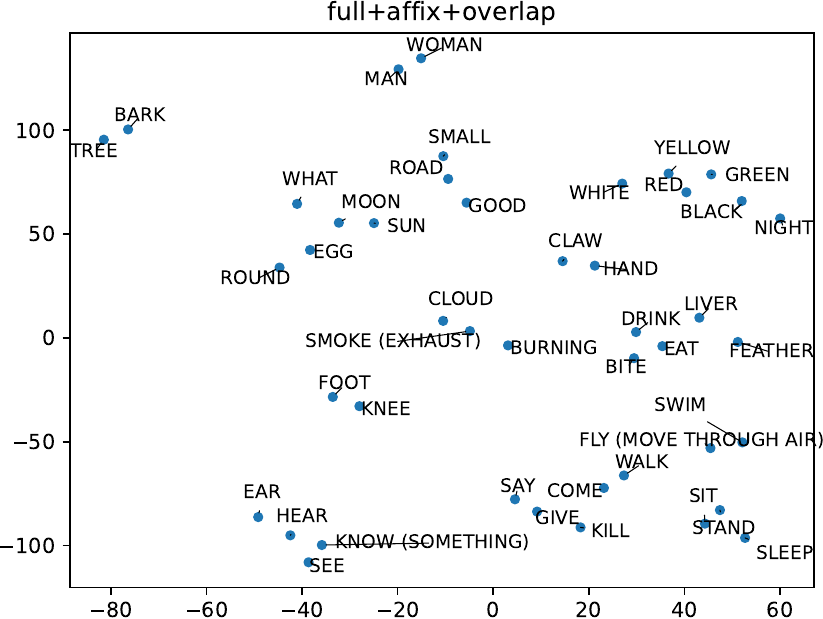} \\
    \end{tabular}}
    \caption{Two-dimensional visualization of embeddings for concepts from the Swadesh-100 list (Swadesh 1955) that are present in all three colexification networks, created using t-SNE.}
    \label{fig:tsne-all}
\end{figure*}

Figure \ref{fig:tsne-all} shows shows two-dimensional visualizations of all different combinations of full and partial colexifications tested in our study for a small list oc concepts taken from the Swadesh list of 100 items \citep{Swadesh1955}.

\newpage
\section{Hyperparameter Choice}

The models discussed in the paper were trained on an Apple M2 Pro CPU with the following hyperparameters:

\begin{table}[!h]
\begin{tabular}{@{}lll@{}}
\toprule
\textbf{Model}    & \textbf{Hyperparameters} & \textbf{Training Time} \\ \midrule
SDNE     & $\textrm{hidden sizes}=(256, 128), \alpha=0.2, \beta=10, $    & 02:00 min            \\
   &  $\textrm{learning rate}=0.001, \textrm{epochs}=10000$.    &             \\
Node2Vec & $n \textrm{(number of random walks per start node)=5},$    & 13:27 min           \\
 & $w\textrm{(length of random walks)}=10, p=1, q=1,$ \\ & $c\textrm{(context window size)}=2, \textrm{test split}=0.2,$     &            \\
 & $w\textrm{(length of random walks)}=10, p=1, q=1,$ \\ & $\textrm{learning rate}=0.001, \textrm{epochs}=1500$.     &            \\
 & Underlying Word2Vec model: SkipGram, no negative sampling. \\
ProNE    &  $\textrm{step}=10, \mu=0.2, \theta=0.5, \textrm{exponent}=0.75$. & 38 ms              \\ \bottomrule
\end{tabular}
\end{table}

\end{document}